\documentclass[letterpaper,10pt,conference]{ieeeconf}
\IEEEoverridecommandlockouts
\usepackage[T1]{fontenc}
\usepackage[utf8]{inputenc}
\usepackage{times,amsmath,amssymb,graphicx,booktabs,tabularx,makecell}
\usepackage[table]{xcolor}
\usepackage{cite,url,capt-of,placeins,float}
\usepackage{microtype,needspace,balance}
\usepackage{eso-pic}
\definecolor{HeaderRule}{HTML}{505054}
\newcommand{\institutionheader}{%
  \AddToShipoutPictureFG*{\AtPageUpperLeft{%
    \put(54,-39){\includegraphics[height=26pt]{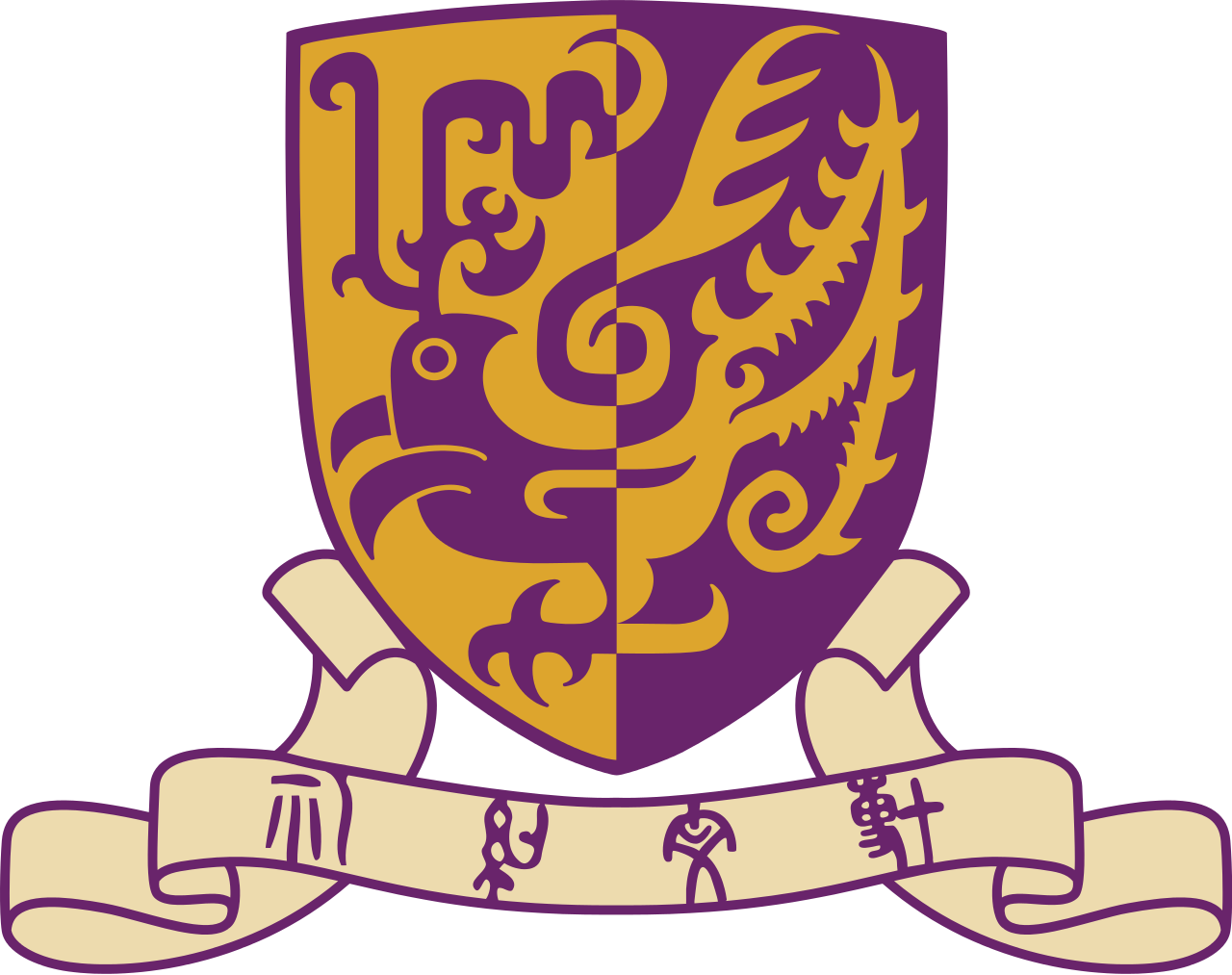}}%
    \put(98,-38){\includegraphics[height=24pt]{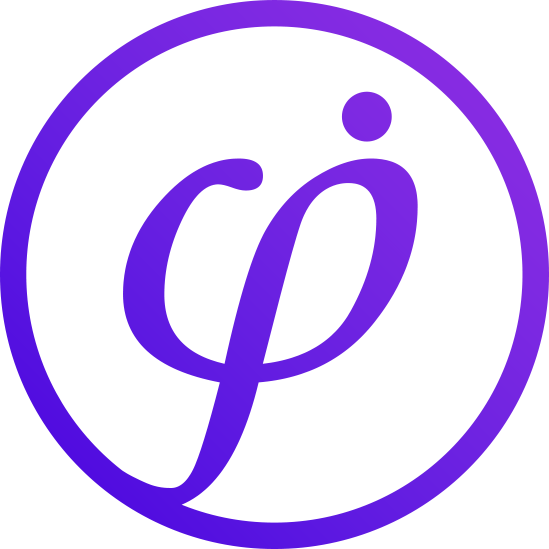}}%
    \put(54,-46){\color{HeaderRule}\rule{\textwidth}{0.4pt}}%
  }}}
\definecolor{NeutralLightGray}{HTML}{E6E8EA}
\newcolumntype{Y}{>{\raggedright\arraybackslash}X}

\newcommand{\ours}{WholeBodyWAM}

\title{WholeBodyWAM: Generalizing Pre-trained\newline
World--Action Priors to Humanoid Loco-Manipulation\newline
via WBC-Grounded Coordination}
\author{Zhuo Li$^{1,4\dagger}$\quad Yiming Yao$^{2,4*}$\quad Jim Tan$^{3,4*}$\quad
Mengjie Jing$^{1,4}$\quad Zhipeng Dong$^{1,4}$\quad Fei Chen$^{1,4\ddagger}$}
\makeatletter
\def\@maketitle{\newpage
  \begin{center}
    {\sffamily\bfseries\fontsize{19}{22}\selectfont\@title\par}
    \vspace{10pt}
    {\normalfont\fontsize{11}{15}\selectfont\@author\par}
    \vspace{8pt}
    {\normalfont\fontsize{10.5}{14}\selectfont
      $^1$The Chinese University of Hong Kong\quad
      $^2$The University of Hong Kong\par
      $^3$Peking University\quad $^4\Phi$-Institute\par}
    \vspace{8pt}
    {\normalfont\fontsize{10}{13}\selectfont
      $^\dagger$Project Lead\qquad $^\ddagger$Corresponding Author\qquad
      $^*$Equal Contribution\par}
  \end{center}}
\makeatother
\newcommand{\teaserblock}{%
  \begin{minipage}{\textwidth}
  \centering
  \includegraphics[width=\textwidth]{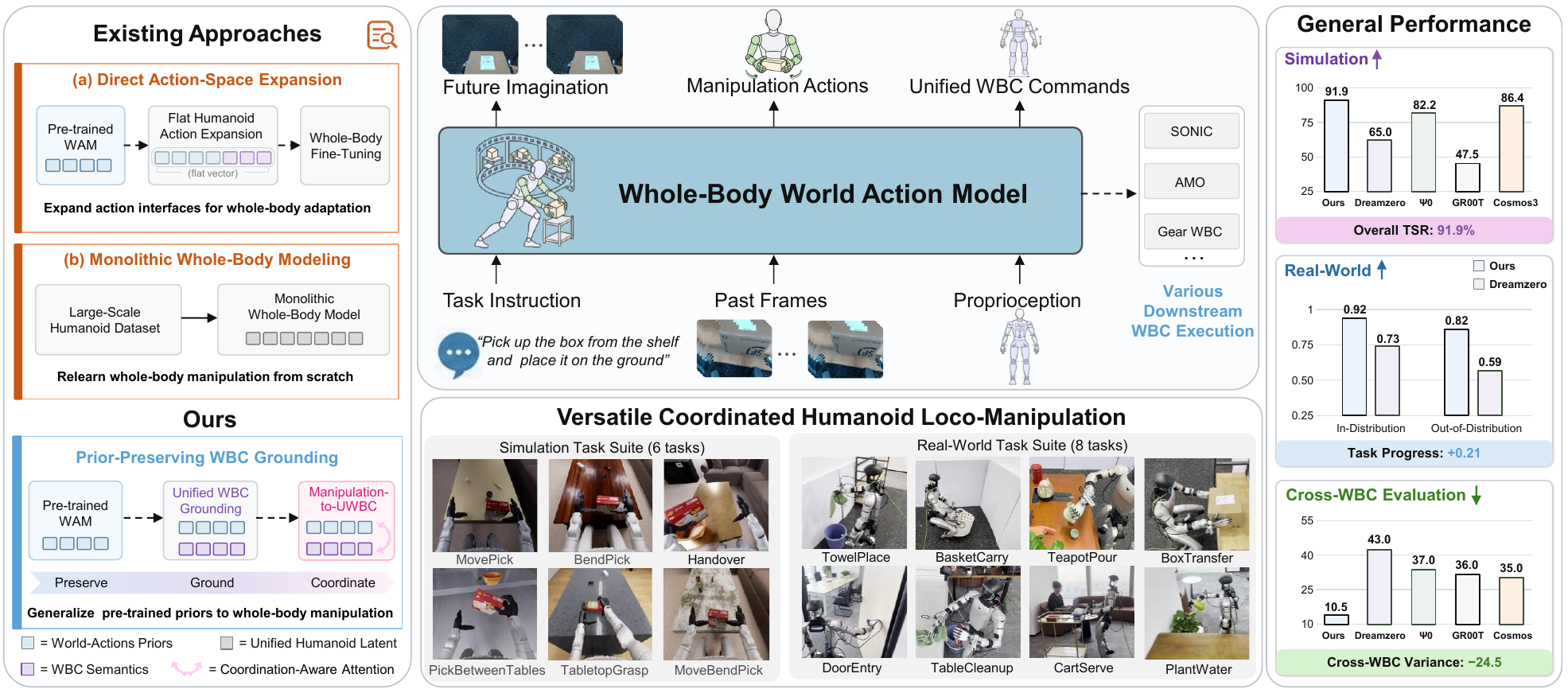}
  \captionof{figure}{\textbf{Overview of WholeBodyWAM.} Unlike existing approaches that expand action interfaces for whole-body adaptation or learn dedicated humanoid world--action mappings from scratch, \ours{} preserves and generalizes the world--action priors of pre-trained WAMs to humanoid loco-manipulation through WBC-grounded coordination. Given language, visual history, and proprioception, it jointly predicts future visual dynamics, manipulation intent, and unified whole-body controller (WBC) commands. Evaluations on six simulation and eight real-world tasks demonstrate higher overall task success rates and lower cross-controller performance variance than the evaluated baselines, supporting effective loco-manipulation across diverse tasks and heterogeneous WBCs.}
  \label{fig:teaser}
  \end{minipage}\par\vspace{1em}}
\IEEEaftertitletext{\teaserblock}

\begin{document}
\institutionheader
\maketitle
\thispagestyle{empty}
\pagestyle{empty}
\bstctlcite{IEEEcontrol}
\begin{abstract}
World Action Models (WAMs) offer a promising approach to general-purpose robot manipulation by jointly modeling visual dynamics and actions. However, most WAM studies focus on tabletop or arm-centric manipulation, while humanoid loco-manipulation remains less explored. To address this gap, we introduce WholeBodyWAM, which jointly predicts future visual dynamics, manipulation actions, and whole-body control intents for generalizable humanoid loco-manipulation. It preserves pre-trained world--action priors while grounding heterogeneous whole-body controller (WBC) semantics and coordinating whole-body behavior. Extensive experiments show that WholeBodyWAM achieves an overall simulation task success rate of 91.9\%, with a 0.23 improvement in real-world out-of-distribution task progress and a 70\% reduction in success-rate variance across WBCs relative to the respective baselines. These results suggest a path toward scalable humanoid whole-body intelligence by extending pre-trained world--action priors through structured WBC grounding and coordination, rather than relearning whole-body behavior from scratch.
Project page: \url{https://wholebodywam.github.io/}.
\end{abstract}

\section{Introduction}
World Action Models (WAMs) jointly model robot actions and future visual dynamics, providing reusable priors for general robot manipulation~\cite{dreamzero,dit4dit,fastwam}. Most existing studies investigate local hand--object interactions in tabletop or arm-centric settings~\cite{cosmospolicy,fastwam,lingbotva,motus}, while wide-range, whole-body loco-manipulation on bipedal humanoids remains underexplored. This gap stems from three coupled challenges:
\textit{Whole-body coordination complexity.} Successful humanoid loco-manipulation depends not only on generating plausible hand--object interactions but also on modeling when and how whole-body motion should coordinate with manipulation.
\textit{Heterogeneous WBC semantics.} Existing WBCs expose command ontologies with different physical meanings~\cite{sonic,amo,hugwbc}, resulting in similar motions requiring different combinations of task-space commands and joint targets. This heterogeneity hinders reusable action semantics and consistent action--dynamics modeling across controllers.
\textit{Humanoid data scarcity.} Despite the availability of large-scale robot data for WAM pretraining, paired humanoid loco-manipulation demonstrations remain scarce~\cite{wholebodyvla,egohumanoid,psi}. Such scarcity constrains direct supervision for modeling the coordination between manipulation intent and the whole-body adjustments required for execution.

Recent studies have begun exploring WAMs for humanoid loco-manipulation~\cite{motionwam,omega,dit4dit}. These approaches fall into two broad categories. \textit{1) Direct action-space expansion:} One straightforward strategy is to expand the action interface of a pre-trained tabletop WAM and fine-tune the model on whole-body demonstrations for humanoid adaptation. DiT4DiT~\cite{dit4dit} demonstrates related fine-tuning of video--action models for humanoid manipulation. Although this strategy preserves transferable world--action priors, it leaves heterogeneous WBC semantics and complex whole-body coordination to be learned implicitly from limited fine-tuning supervision. Consequently, the model may retain high-level manipulation intent yet produce poorly grounded WBC commands and temporally inconsistent whole-body behavior, limiting learning efficiency and generalization on coordination-intensive tasks.
\textit{2) Monolithic whole-body modeling:} An alternative approach learns humanoid-native whole-body WAMs that unify locomotion and manipulation within controller-compatible latents~\cite{motionwam,omega}. Although these models support coherent whole-body motion, such approaches can entangle policy prediction with controller-specific conventions (e.g., SONIC motion latents) and remain dependent on costly large-scale whole-body supervision.

These limitations raise a central question: \textit{How can pre-trained world--action priors be generalized to humanoid loco-manipulation without learning whole-body behavior from scratch?}

Our key insight is that pre-trained WAMs already capture rich manipulation priors, but what is fundamentally missing for humanoid loco-manipulation is structured grounding to heterogeneous WBCs that regulates when and how whole-body behavior should coordinate. Based on this insight, we propose \ours{}, a WBC-grounded world action model that generalizes learned world--action priors to coordinated humanoid loco-manipulation through three complementary designs.

First, a \textit{structured action representation} retains the manipulation pathway of the pre-trained WAMs and introduces a distinct whole-body control stream, with visual dynamics and both action streams jointly generated by a shared Diffusion Transformer (DiT). Second, the \textit{Unified Whole-Body Controller Interface (UWBC)} assigns consistent physical meanings to shared commands while accommodating controller-specific extensions. Third, \textit{Coordination-Aware Self-Attention (CASA)} uses task-directional arm manipulability to modulate an attention bias that strengthens manipulation-to-UWBC information flow when greater whole-body coordination is indicated.

Together, these designs preserve world--action priors of the pre-trained WAMs, ground heterogeneous WBC commands, and coordinate whole-body behavior for generalizable humanoid loco-manipulation.
In summary, our contributions are threefold:
\begin{itemize}[\setlength{\topsep}{2pt plus 0.5pt}\setlength{\partopsep}{0pt}\setlength{\parsep}{0pt}\setlength{\itemsep}{0pt plus 0.5pt}\setlength{\parskip}{0pt}]
 \item We present \ours{}, a world action model that generalizes pre-trained world--action priors to humanoid loco-manipulation through coordinated WBC grounding.
 \item We develop UWBC as a shared semantic interface for heterogeneous WBCs and CASA for manipulation-informed whole-body coordination.
 \item Experiments on six simulation and eight real-world tasks demonstrate improved task performance, OOD generalization, and cross-WBC robustness.
\end{itemize}

\section{Related Work}
\subsection{World Action Models}
World Action Models couple visual dynamics with action generation to enable generalizable robot manipulation~\cite{dreamzero,cosmospolicy,fastwam,dit4dit,motionwam,omega}. DreamZero~\cite{dreamzero} jointly models future video and actions. Cosmos Policy~\cite{cosmospolicy} encodes actions, future states, and values as latent frames within a pre-trained video diffusion model, while Fast-WAM~\cite{fastwam} co-trains video and action prediction but removes future-video tokens at inference. Recent work has begun to extend this paradigm from tabletop manipulation toward humanoid loco-manipulation. DiT4DiT~\cite{dit4dit} applies video--action modeling to bimanual manipulation on a humanoid platform. MotionWAM~\cite{motionwam} predicts a unified humanoid motion latent, while $\omega$-0~\cite{omega} generates controller-compatible whole-body action latents. In contrast to expanding action interfaces or relearning humanoid world--action mappings, \ours{} adapts pre-trained WAMs by learning the WBC grounding and coordination needed for humanoid execution. This allows limited humanoid supervision to extend a reusable world--action prior to whole-body configurations.

\subsection{Humanoid Loco-Manipulation}
Learning-based humanoid loco-manipulation has advanced from whole-body motor control to task-level policy learning. OmniH2O~\cite{omnih2o}, HOVER~\cite{hover}, and HugWBC~\cite{hugwbc} provide robust tracking and versatile whole-body control interfaces for externally specified commands. Building on these foundations, VLA-based methods such as WholeBodyVLA~\cite{wholebodyvla}, $\Psi_0$~\cite{psi}, HEX~\cite{hex}, and OpenHLM~\cite{openhlm} learn autonomous whole-body policies from vision--language priors and human or cross-embodiment supervision. WAM-based MotionWAM~\cite{motionwam} and $\omega$-0~\cite{omega} further incorporate predictive world modeling into whole-body action generation. These policies generally use action representations tailored to specific downstream WBC interfaces. In contrast, \ours{} explicitly learns a shared semantic interface over heterogeneous WBC commands through UWBC. This enables reusable WBC grounding priors to be adapted across different downstream controllers.

\begin{figure*}[t]
 \centering
 \includegraphics[width=.96\textwidth]{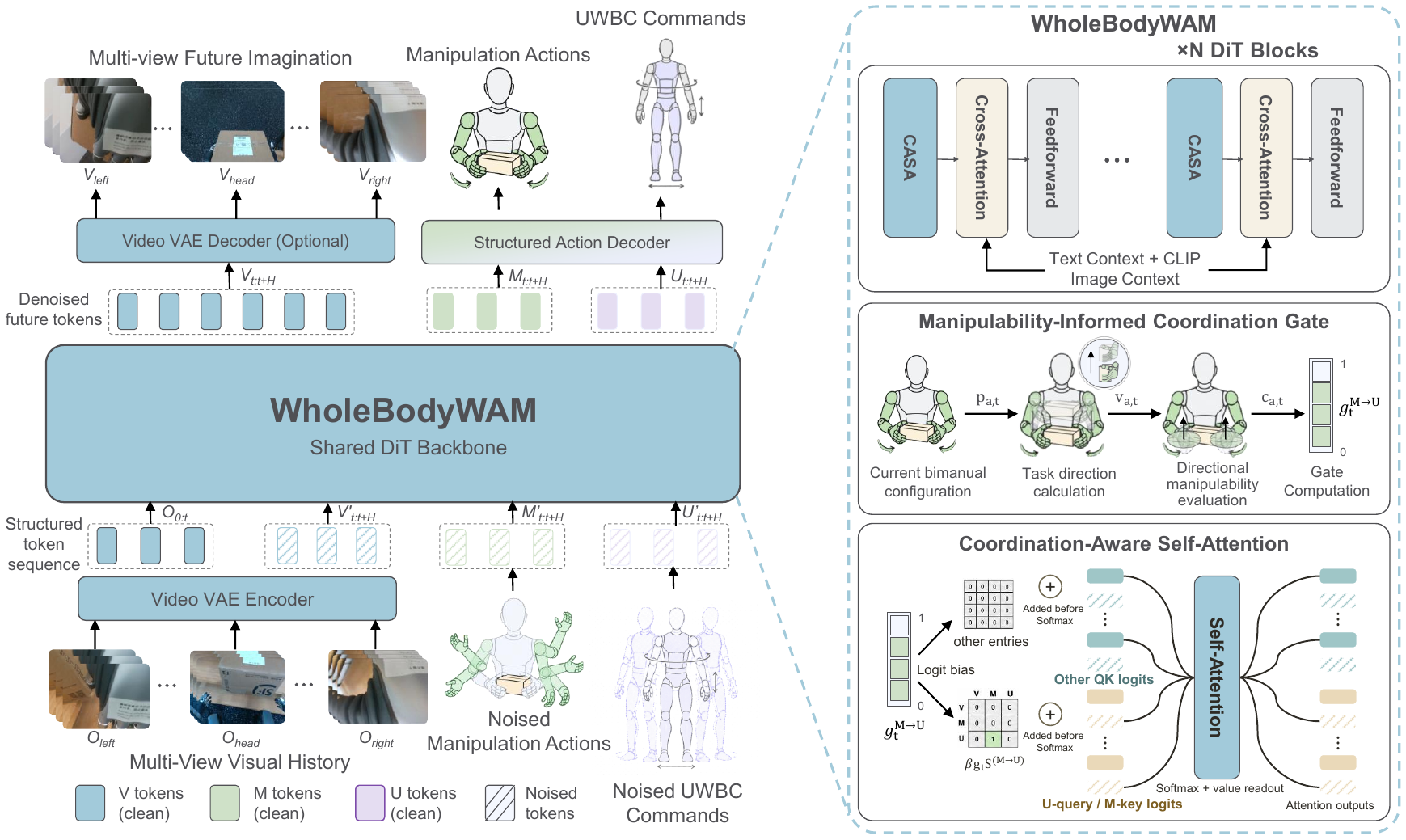}
 \caption{\textbf{Model Architecture.} \textbf{Left:} A shared Diffusion Transformer jointly generates future visual dynamics, manipulation actions, and UWBC commands from a structured token sequence conditioned on visual history and task context. \textbf{Right:} CASA adaptively strengthens manipulation-to-UWBC attention based on reductions in task-directional arm manipulability, allowing manipulation intent to guide whole-body coordination.}
 \label{fig:architecture}
\end{figure*}

\section{Method}
\subsection{Problem Formulation}
\ours{} formulates humanoid loco-manipulation as joint prediction of future world dynamics and structured whole-body actions under the task context:
\begin{equation}
 p_\theta(\mathbf v_{t:t+H},\mathbf m_{t:t+H},\mathbf u_{t:t+H}\mid\mathbf c_t).
 \label{eq:joint}
\end{equation}
The context $\mathbf c_t$ comprises visual history $\mathbf o_{0:t}$, current robot state $\mathbf s_t$, and language instruction $\ell$. Here $H$ is the prediction horizon, and $\mathbf v_{t:t+H}$ denotes future visual dynamics. The manipulation stream $\mathbf m_{t:t+H}$ specifies arm joint targets and finger joint angles for task-level manipulation. The stream $\mathbf u_{t:t+H}$ carries whole-body commands expressed through the UWBC interface.
Following joint world--action modeling~\cite{dreamzero}, the conditional distribution in Eq.~\eqref{eq:joint} can be factorized as:
\begin{equation}
\begin{aligned}
 &p_\theta(\mathbf v_{t:t+H},\mathbf m_{t:t+H},\mathbf u_{t:t+H}\mid\mathbf c_t)\\
 &\quad=p_\theta^v(\mathbf v_{t:t+H}\mid\mathbf c_t)\,
 p_\theta^a(\mathbf m_{t:t+H},\mathbf u_{t:t+H}\mid
 \mathbf v_{t:t+H},\mathbf c_t).
\end{aligned}
\label{eq:factorization}
\end{equation}
The world factor $p_\theta^v$ models future scene evolution conditioned on the task context, while the action factor $p_\theta^a$ jointly models manipulation actions and UWBC commands conditioned on both the task context and the corresponding future dynamics. This factorization describes the dependency structure of the joint world--action distribution rather than a sequential video-then-action inference procedure. The visual, manipulation, and UWBC streams are instantiated and optimized jointly within a shared generative backbone, as detailed in the following subsection.

\subsection{Model Architecture}
As illustrated in Fig.~\ref{fig:architecture}, \ours{} builds upon the pre-trained video Diffusion Transformer (DiT) backbone~\cite{dreamzero}, which provides a shared generative model for visual dynamics and robot actions. We extend the existing world--action token sequence with a structured UWBC stream, allowing visual, manipulation, and whole-body motion to be jointly modeled.
The pre-trained T5 text encoder~\cite{t5} provides instruction features through cross-attention. The pre-trained Wan video VAE~\cite{wan} compresses observations into latent video tokens, while a lightweight state encoder maps the current robot state to proprioceptive conditioning tokens.
Future visual tokens $\mathbf v$, manipulation tokens $\mathbf m$, and UWBC tokens form three typed streams that share a prediction block at each future step and are jointly processed by the DiT. The manipulation stream retains the visual--manipulation pathway inherited from the pre-trained WAM~\cite{dreamzero}, while distinct UWBC tokens make controller-facing whole-body behavior explicitly addressable.
Following the joint flow-matching formulation~\cite{dreamzero}, \ours{} minimizes a modality-weighted objective~\cite{flow}:
\begin{equation}
\mathcal L(\theta)=\mathbb E\!\left[
\sum_k\lambda_k w_k(\tau)
\left\|f_\theta^k(\mathbf x_\tau,\tau,\mathbf c_t)
-\dot{\mathbf x}^{k,*}\right\|_k^2\right].
\label{eq:jfm}
\end{equation}
Here $\mathbf x_\tau$ denotes the noisy joint block at flow time $\tau\in[0,1]$, and $\dot{\mathbf x}^{k,*}$ is the target flow velocity for stream $k\in\{v,m,u\}$. The coefficient $\lambda_k$ balances the stream losses, while $w_k(\tau)$ is a loss weight that varies with flow time.

\subsection{Unified WBC Interface}
Existing whole-body controllers expose heterogeneous command interfaces spanning task-space objectives, discrete control modes, and joint-space references with different physical semantics~\cite{sonic,amo}. Naively concatenating these quantities into a flat action vector obscures their control semantics and tightly couples the learned policy to a specific controller interface. We therefore introduce the Unified Whole-Body Controller Interface (UWBC) to organize heterogeneous whole-body commands according to shared physical semantics while supporting controller-specific extensions.
As shown in Table~\ref{tab:uwbc_slot_specification}, UWBC defines shared continuous command fields with consistent physical meanings across controllers. Task-space whole-body commands, together with lower-body and waist joint positions and velocities, form a 46-D block of shared continuous commands. Individual WBCs may require additional controller-specific commands, which are accommodated by ten residual slots at indices $[46,56)$. The command vector expressed through UWBC is defined as
\begin{equation}
\begin{aligned}
 \mathbf u_t&=[\mathbf u_t^{\mathrm{sh}},\mathbf u_t^r]\in\mathbb R^{56},\\
 \mathbf u_t^{\mathrm{sh}}
 &=[\mathbf u_t^{\mathrm{task}},\mathbf q_t^{\mathrm{lower}},
 \dot{\mathbf q}_t^{\mathrm{lower}}]\in\mathbb R^{46}.
\end{aligned}
\label{eq:uwbc}
\end{equation}
Here $\mathbf u_t^{\mathrm{sh}}$ collects the shared slots, $\mathbf u_t^{\mathrm{task}}\in\mathbb R^{16}$ contains task-space commands, and $\mathbf q_t^{\mathrm{lower}},\dot{\mathbf q}_t^{\mathrm{lower}}\in\mathbb R^{15}$ specify desired lower-body and waist joint positions and velocities. The residual slots $\mathbf u_t^r\in\mathbb R^{10}$ encode controller-specific commands.
For a downstream controller, UWBC fields are activated according to their registered physical semantics and controller capability profile, while unsupported or undefined fields are masked.

\begin{figure}[t]
 \centering
 \includegraphics[width=.78\columnwidth]{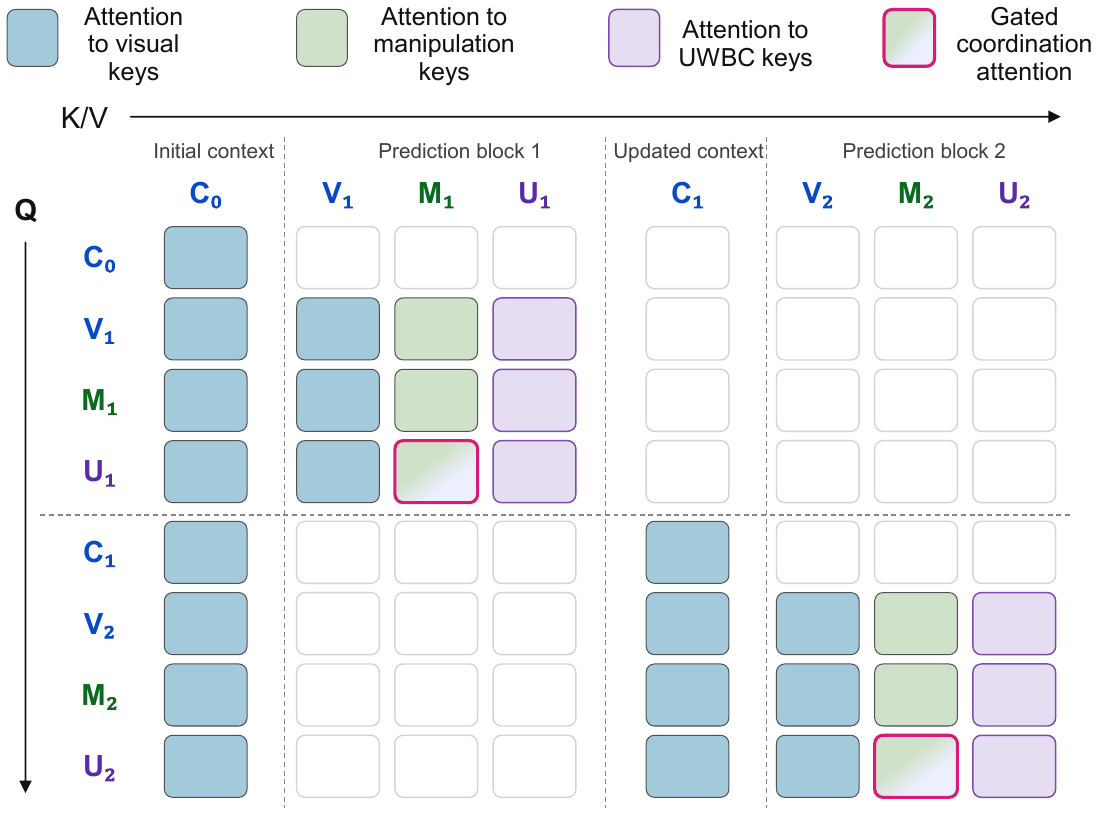}
 \caption{\textbf{Blockwise-Causal Attention with Gated Coordination.} Native self-attention visibility is preserved across visual, manipulation, and UWBC streams, while a gated coordination bias selectively enhances manipulation-to-UWBC attention within each prediction block.}
 \label{fig:attention}
\end{figure}

\subsection{Coordination-Aware Self-Attention}
Coordination in humanoid loco-manipulation is inherently directional and state dependent, requiring compensatory body motions when arm manipulability along the task direction becomes insufficient. We therefore introduce a \textbf{coordination-aware self-attention mechanism} that selectively strengthens manipulation-to-UWBC information flow.

\suppressfloats[t]
\begin{table}[t]
  \centering
  \definecolor{UWBCCharcoal}{HTML}{505054}
  \color{UWBCCharcoal}
  \arrayrulecolor{UWBCCharcoal}
  \caption{UWBC interface slot specification: shared physical-semantic fields and controller-specific residual slots.}
  \label{tab:uwbc_slot_specification}
  \footnotesize
  \setlength{\tabcolsep}{3.0pt}
  \renewcommand{\arraystretch}{1.0}
  \begin{tabularx}{\columnwidth}{@{}ccY@{}}
    \toprule
    \makecell[c]{\textbf{Index}\\\textbf{Range}}
    & \makecell[c]{\textbf{Element}\\\textbf{Index}}
    & \textbf{Mapped Physical Semantics} \\
    \midrule
    \rowcolor{NeutralLightGray}
    \multicolumn{3}{c}{\textbf{Shared slots} $[0,46)$ (46-D)} \\
    \texttt{[0,3)} & 0--2 & Desired base linear velocity $(v_x,v_y,v_z)$ \\
    \texttt{[3,6)} & 3--5 & Desired base angular velocity $(\omega_x,\omega_y,\omega_z)$ \\
    \texttt{[6,8)} & 6--7 & Desired heading $(\sin\psi,\cos\psi)$ \\
    \texttt{[8,10)} & 8--9 & Desired pelvis planar position $(x,y)$ \\
    \texttt{[10,11)} & 10 & Desired pelvis height \\
    \texttt{[11,13)} & 11--12 & Desired pelvis tilt (roll/pitch) \\
    \texttt{[13,16)} & 13--15 & Desired torso-to-pelvis orientation (roll/pitch/yaw) \\
    \cmidrule(lr){1-3}
    \texttt{[16,31)} & 16--30 & Desired lower-body/waist joint positions \\
    \texttt{[31,46)} & 31--45 & Desired lower-body/waist joint velocities \\
    \midrule
    \rowcolor{NeutralLightGray}
    \multicolumn{3}{c}{\textbf{Residual slots} $[46,56)$ (10-D)} \\
    \texttt{[46,56)} & 46--55 & Registered controller-specific commands, including discrete control modes \\
    \bottomrule
  \end{tabularx}
  \arrayrulecolor{black}
\end{table}

\textit{Gated attention logit bias.} The native self-attention logits are modified as follows:
\begin{equation}
\mathbf L'_t=\frac{\mathbf Q\mathbf K^\top}{\sqrt d}
 +\mathbf M^{\mathrm{nat}}
 +\beta g_t\mathbf S^{M\rightarrow U}.
\label{eq:logit}
\end{equation}
Here $\mathbf Q$ and $\mathbf K$ denote the query and key matrices, $d$ denotes the head dimension, and $\mathbf M^{\mathrm{nat}}$ represents the native self-attention mask. The coordination gate $g_t\in[0,1]$ scales a fixed bias of strength $\beta>0$. As illustrated in Fig.~\ref{fig:attention}, the selector $S_{ij}^{M\rightarrow U}=1$ only when $i$ is a UWBC query and $j$ is a manipulation key in the same prediction block.

\textit{Coordination gate computation.} The gate $g_t$ evaluates reduced task-directional manipulability in the current arm configurations to determine whether whole-body coordination is needed. First, for each arm $a\in\{L,R\}$, the task direction is calculated from consecutive observed end-effector positions during motion:
\begin{equation}
 \mathbf v_{a,t}
 =\frac{\mathbf p_{a,t}-\mathbf p_{a,t-1}}{\Delta t},
 \qquad
 \hat{\mathbf d}_{a,t}
 =\frac{\mathbf v_{a,t}}{\|\mathbf v_{a,t}\|_2}.
 \label{eq:task_direction}
\end{equation}
Here $\mathbf p_{a,t}$ denotes the observed end-effector position, $\Delta t$ denotes the observation interval, and $\mathbf v_{a,t}$ denotes the observed end-effector velocity. The unit vector $\hat{\mathbf d}_{a,t}$ represents the estimated task direction.
The manipulability of each arm configuration along this task direction is then calculated as
\begin{equation}
 c_{a,t}
 =\left[\hat{\mathbf d}_{a,t}^{\top}
 \bigl(C_a^{v,\lambda}(\mathbf q^{\mathrm{obs}}_{a,t})\bigr)^{-1}
 \hat{\mathbf d}_{a,t}\right]^{-1/2}.
 \label{eq:directional_capability}
\end{equation}
Here $C_a^{v,\lambda}\in\mathbb R^{3\times3}$ denotes the damped manipulability matrix incorporating joint speed limits, and $\mathbf q^{\mathrm{obs}}_{a,t}$ denotes the observed arm joint configuration.
We construct a voxelized manipulability map offline following~\cite{vahrenkamp2012}. At runtime, the reference $c^{\mathrm{ref}}_{a,t}$ is the maximum sampled radial capability along the estimated task direction among configurations in the current end-effector pose voxel. The gate for each arm is then calculated from the relative reduction in manipulability:
\begin{equation}
 g_{a,t}=1-\operatorname{clip}\!\left(
 \frac{c_{a,t}}{c^{\mathrm{ref}}_{a,t}+\epsilon_m},0,1\right),
 \label{eq:arm_gate}
\end{equation}
where $\epsilon_m>0$ is a small constant that prevents division by zero. Finally, the global gate for both arms takes the larger relative reduction:
\begin{equation}
 g_t^{M\rightarrow U}=\max(g_{L,t},g_{R,t}).
 \label{eq:gate}
\end{equation}
This ensures that reduced task-directional manipulability in either active arm can trigger manipulation-to-UWBC coordination.

\setcounter{topnumber}{1}
\begin{figure}[t]
\centering
 \includegraphics[width=\columnwidth]{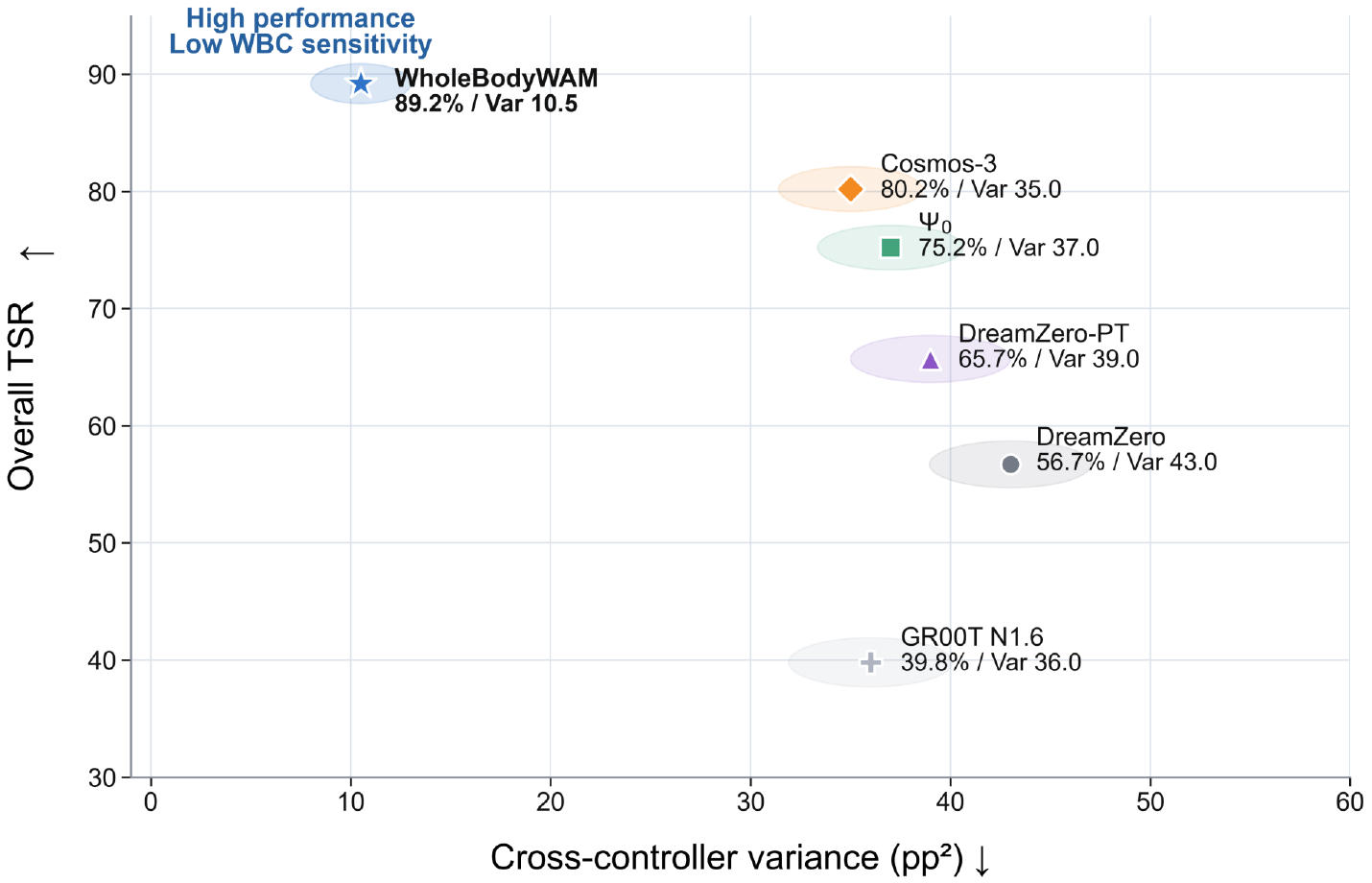}
 \caption{\textbf{Cross-WBC evaluation.} Mean TSR and population variance across SONIC, AMO, and GEAR WBC, with separate fine-tuning for each controller. \ours{} achieves the highest mean TSR (89.2\%) and lowest variance (10.5 pp$^2$).}
 \label{fig:controllers}
\end{figure}

\section{Experiments}
We conduct comprehensive experiments to investigate the following questions: (1) Does \ours{} improve task success over WAM and VLA baselines? (2) Does UWBC improve stability and portability across heterogeneous WBCs? (3) Can manipulation priors generalize to new whole-body coordination in OOD settings? (4) How does each mechanism contribute to the performance of \ours{}?

\subsection{Experimental Setup}

\begin{figure}[t]
\centering
 \includegraphics[width=\columnwidth]{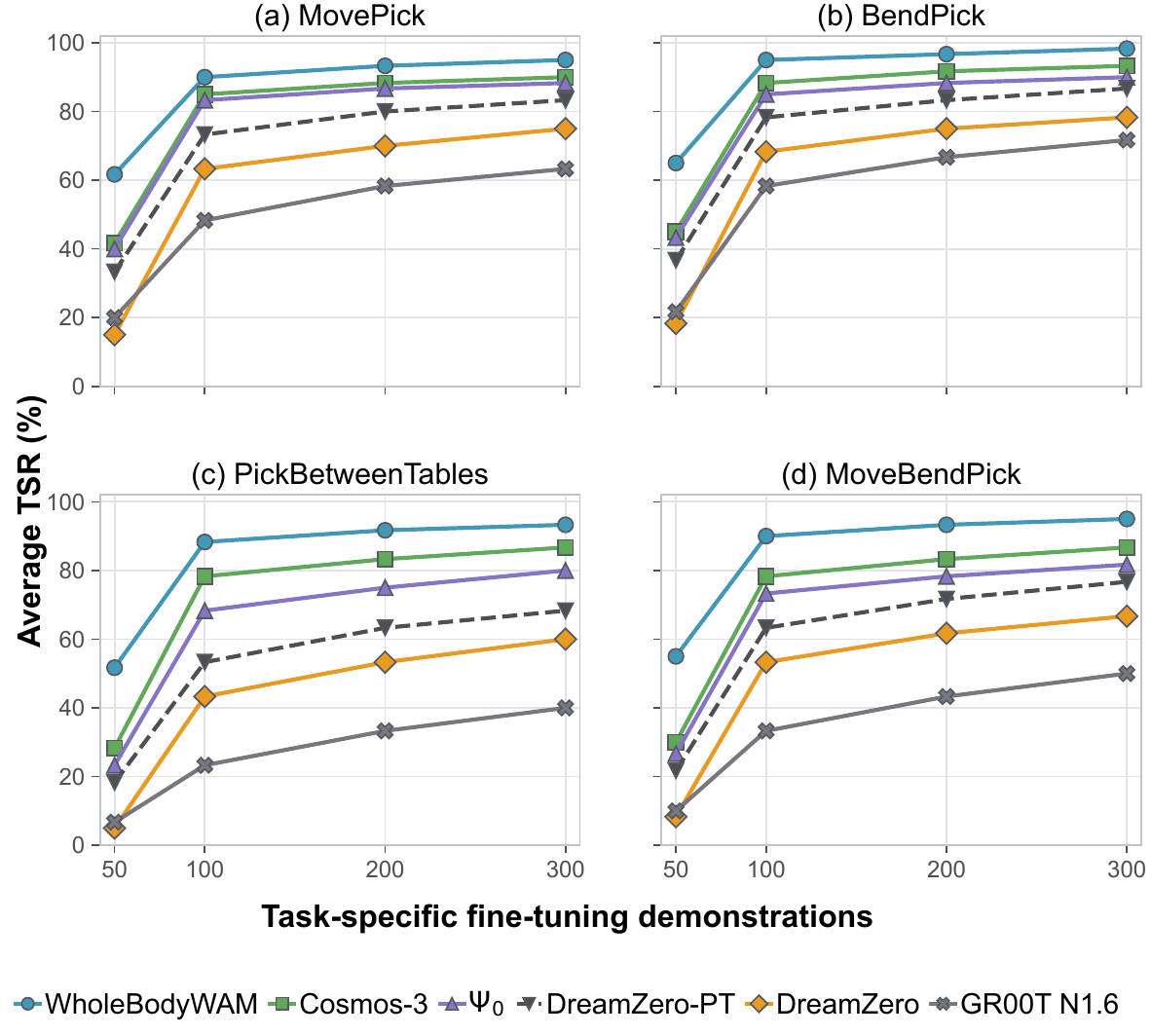}
 \caption{\textbf{Downstream fine-tuning efficiency.} We compare downstream fine-tuning efficiency across methods. \ours{} achieves its largest advantage with few task-specific demonstrations, consistent with effective reuse of pre-trained world--action priors through coordinated WBC grounding.}
 \label{fig:scaling}
\end{figure}
\textit{Training and implementation.} \ours{} is initialized from the pre-trained 14B WAM~\cite{dreamzero}, retaining its world--action backbone. We collect 15K whole-body loco-manipulation demonstrations in SIMPLE~\cite{simple} using a PICO 4 Ultra headset and wrist trackers. Each demonstration is converted into a SMPL motion sequence~\cite{smpl} and retargeted to the G1 embodiment using GMR~\cite{gmr}, yielding desired robot motion references. We then calculate supervision for each UWBC shared slot from the robot motion references according to its predefined physical semantics, coordinate convention, and unit. After temporal alignment, the UWBC targets are incorporated into the post-training dataset. We post-train \ours{} on this dataset using LoRA~\cite{lora}, with rank $r=4$ and scaling parameter $\alpha=4$. The model is optimized for 50,000 steps with a learning rate of $10^{-5}$ and a batch size of 1 per GPU. At inference, we use 16 denoising steps with a second-order UniPC flow-matching sampler and a fixed noise-schedule shift of 5.

\begin{table*}[t]
  \centering
  \caption{Simulation task success rate (\%) with SONIC. Each task entry reports L0/L1/L2 results under the SIMPLE protocol.}
  \label{tab:simulation_quantitative_ablation}
  \setlength{\tabcolsep}{3.2pt}
  \renewcommand{\arraystretch}{1.0}
  \footnotesize
  {
    \begin{tabular*}{\textwidth}{@{\extracolsep{\fill}}lccccccc@{}}
        \toprule
        \textbf{Method}
        & \makecell[c]{\textbf{MovePick}}
        & \makecell[c]{\textbf{BendPick}}
        & \makecell[c]{\textbf{Handover}}
        & \makecell[c]{\textbf{PickBetween}\\\textbf{Tables}}
        & \makecell[c]{\textbf{Tabletop}\\\textbf{Grasp}}
        & \makecell[c]{\textbf{MoveBend}\\\textbf{Pick}}
        & \textbf{Overall} \\
        \midrule
        \rowcolor{NeutralLightGray}
        \multicolumn{8}{c}{\textit{WAM and VLA baselines}} \\
        DreamZero
        & 70/65/55 & 75/70/60 & 85/80/70
        & 50/45/35 & 90/85/75 & 60/55/45
        & 65.0 \\
        DreamZero-PT
        & 80/75/65 & 85/80/70 & 90/85/80
        & 60/55/45 & 95/90/80 & 70/65/55
        & 73.6 \\
        Cosmos-3
        & 90/85/80 & 95/90/80 & \textbf{100}/\textbf{95}/\textbf{90}
        & 85/80/70 & \textbf{100}/\textbf{95}/85 & 85/80/70
        & 86.4 \\
        GR00T N1.6
        & 55/50/40 & 65/60/50 & 60/55/45
        & 30/25/15 & 75/70/60 & 40/35/25
        & 47.5 \\
        $\Psi_{0}$
        & 90/85/75 & 90/85/80 & 95/90/80
        & 75/70/60 & \textbf{100}/\textbf{95}/\textbf{90} & 80/75/65
        & 82.2 \\
        \midrule
        WholeBodyWAM
        & \textbf{95}/\textbf{90}/\textbf{85} & \textbf{100}/\textbf{95}/\textbf{90} & \textbf{100}/\textbf{95}/\textbf{90}
        & \textbf{95}/\textbf{90}/\textbf{80} & \textbf{100}/\textbf{95}/85 & \textbf{95}/\textbf{90}/\textbf{85}
        & \textbf{91.9} \\
        \midrule
        \rowcolor{NeutralLightGray}
        \multicolumn{8}{c}{\textit{Ablations}} \\
        w/o CASA
        & 90/85/80 & 95/90/85 & \textbf{100}/\textbf{95}/85
        & 85/80/75 & 95/90/85 & 90/85/75
        & 86.9 \\
        w/o UWBC
        & 90/85/75 & 95/90/80 & 95/90/85
        & 85/80/70 & 95/90/85 & 85/80/70
        & 84.7 \\
        w/o SAF
        & 85/80/70 & 90/85/75 & 95/90/80
        & 80/75/65 & 95/90/80 & 80/75/65
        & 80.8 \\
        \bottomrule
    \end{tabular*}%
  }
\end{table*}

\begin{figure*}[t]
 \centering
 \includegraphics[width=\textwidth]{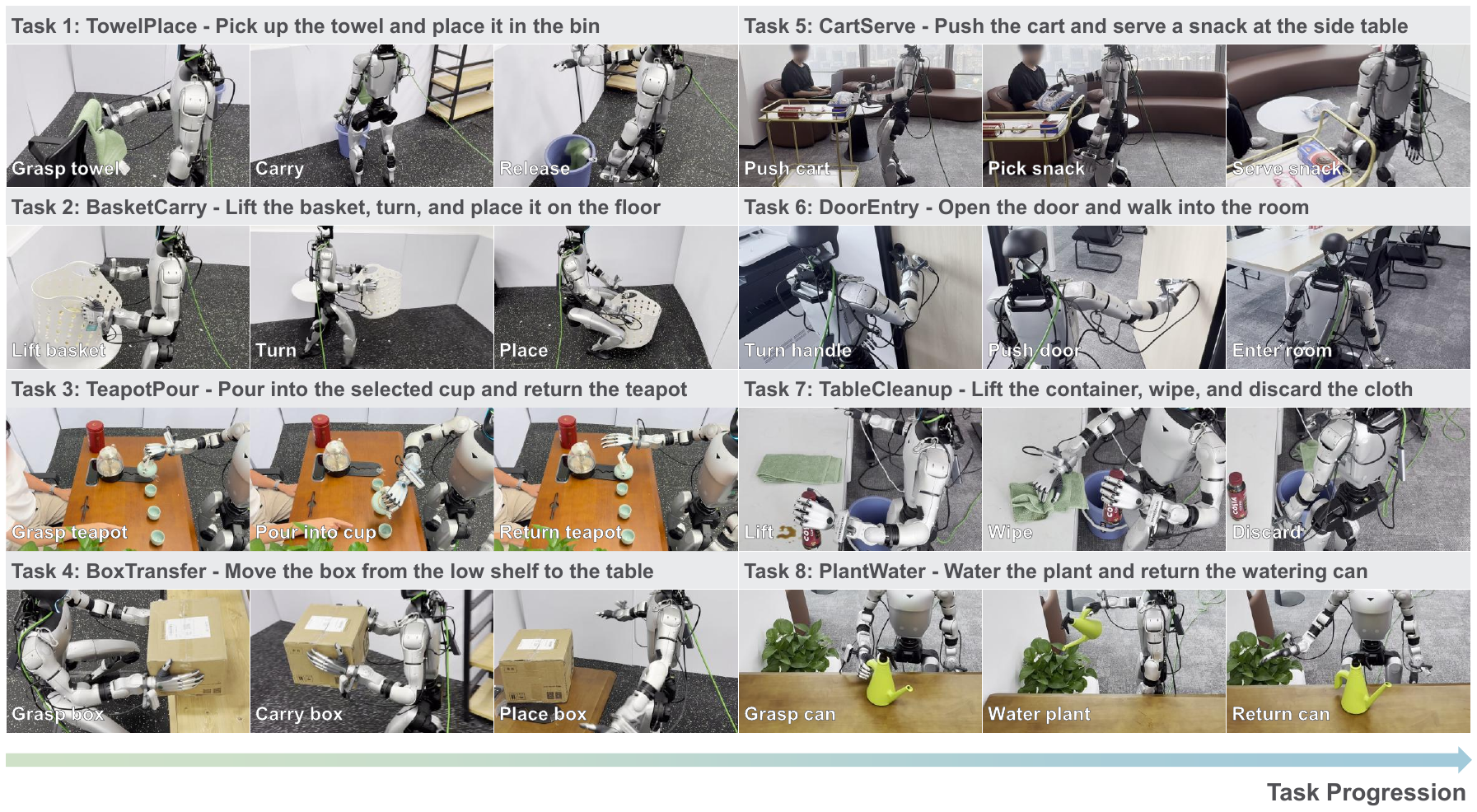}
 \caption{\textbf{Real-world qualitative results.} We show \ours{} performing eight real-world tasks spanning manipulation-dominant and coordination-intensive behaviors. Task instructions appear above each sequence, with frames ordered from left to right. Lower-left labels identify task stages associated with Task Progress (TP) in Fig.~\ref{fig:progress}.}
 \label{fig:real_tasks}
\end{figure*}


\textit{Platforms.} We conduct all real-world experiments on a Unitree G1 humanoid equipped with two BrainCo Revo 2 dexterous hands. Visual observations are captured by one head-mounted and two wrist-mounted Intel RealSense D435i RGB cameras. For simulation, we evaluate humanoid loco-manipulation on the SIMPLE benchmark~\cite{simple}.

\textit{Baselines.} (1) \textit{DreamZero}~\cite{dreamzero} jointly models video and actions for generalizable bimanual manipulation. To adapt its bimanual checkpoint to humanoid tasks, we resize the pre-trained action head to match the G1 embodiment before downstream task fine-tuning. (2) \textit{DreamZero-PT}~\cite{dreamzero} follows the same model and humanoid adaptation setup as DreamZero, with additional post-training on the same dataset as \ours{} before downstream task fine-tuning. (3) \textit{Cosmos-3}~\cite{cosmos} unifies world modeling and action generation, with demonstrated generalization in tabletop manipulation. For humanoid adaptation, we use the Cosmos3-Nano baseline from SIMPLE with a resized humanoid action head and fine-tune it on downstream tasks. (4) \textit{$\Psi_0$}~\cite{psi} is a native whole-body VLA that learns humanoid loco-manipulation from egocentric human videos and humanoid demonstrations. We directly fine-tune its pre-trained model on downstream task demonstrations. (5) \textit{GR00T N1.6}~\cite{groot} is a humanoid VLA pre-trained on diverse robot data. We adapt it through direct fine-tuning on downstream task demonstrations.

\begin{table*}[t]
  \centering
  \caption{Real-world task success rate (\%) under in-distribution (ID) and out-of-distribution (OOD) conditions.}
  \label{tab:real_world_id_ood_tsr}
  \setlength{\tabcolsep}{2.7pt}
  \renewcommand{\arraystretch}{1.0}
  \footnotesize
  {
    \begin{tabular*}{\textwidth}{@{\extracolsep{\fill}}lccccccccc@{}}
        \toprule
        \textbf{Method}
        & \makecell[c]{\textbf{Towel}\\\textbf{Place}}
        & \makecell[c]{\textbf{Basket}\\\textbf{Carry}}
        & \makecell[c]{\textbf{Teapot}\\\textbf{Pour}}
        & \makecell[c]{\textbf{Box}\\\textbf{Transfer}}
        & \makecell[c]{\textbf{Cart}\\\textbf{Serve}}
        & \makecell[c]{\textbf{Door}\\\textbf{Entry}}
        & \makecell[c]{\textbf{Table}\\\textbf{Cleanup}}
        & \makecell[c]{\textbf{Plant}\\\textbf{Water}}
        & \textbf{Overall} \\
        \midrule
        \rowcolor{NeutralLightGray}
        \multicolumn{10}{c}{\textit{In-Distribution Evaluation}} \\
        WholeBodyWAM
        & \textbf{85} & \textbf{75} & \textbf{85} & \textbf{75}
        & \textbf{80} & \textbf{80} & \textbf{90} & \textbf{80}
        & \textbf{81.3} \\
        DreamZero
        & 65 & 40 & 75 & 35
        & 50 & 45 & 85 & 65
        & 57.5 \\
        \midrule
        \rowcolor{NeutralLightGray}
        \multicolumn{10}{c}{\textit{Out-of-Distribution Evaluation}} \\
        WholeBodyWAM
        & \textbf{75} & \textbf{55} & \textbf{80} & \textbf{60}
        & \textbf{65} & \textbf{70} & \textbf{80} & \textbf{65}
        & \textbf{68.8} \\
        DreamZero
        & 50 & 15 & 65 & 20
        & 30 & 25 & 70 & 45
        & 40.0 \\
        \bottomrule
    \end{tabular*}%
  }
\end{table*}

\begin{figure}[t]
 \centering
 \includegraphics[width=\columnwidth]{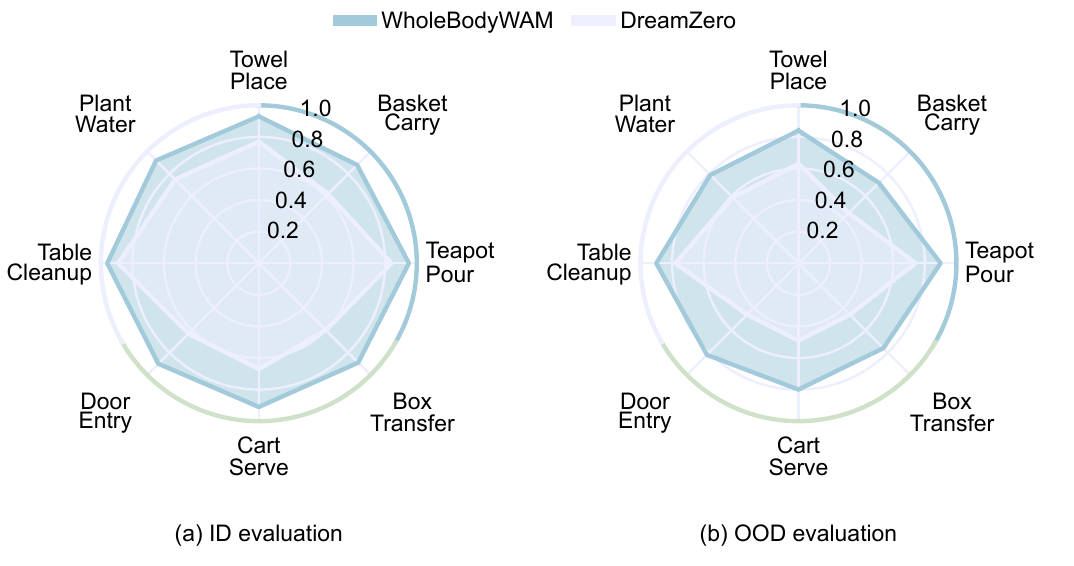}
 \caption{\textbf{Task progress under ID (left) and OOD (right).} \ours{} maintains strong progress across eight tasks spanning manipulation-dominant and coordination-intensive behaviors.}
 \label{fig:progress}
\end{figure}

\begin{figure}[t]
 \centering
 \includegraphics[width=\columnwidth]{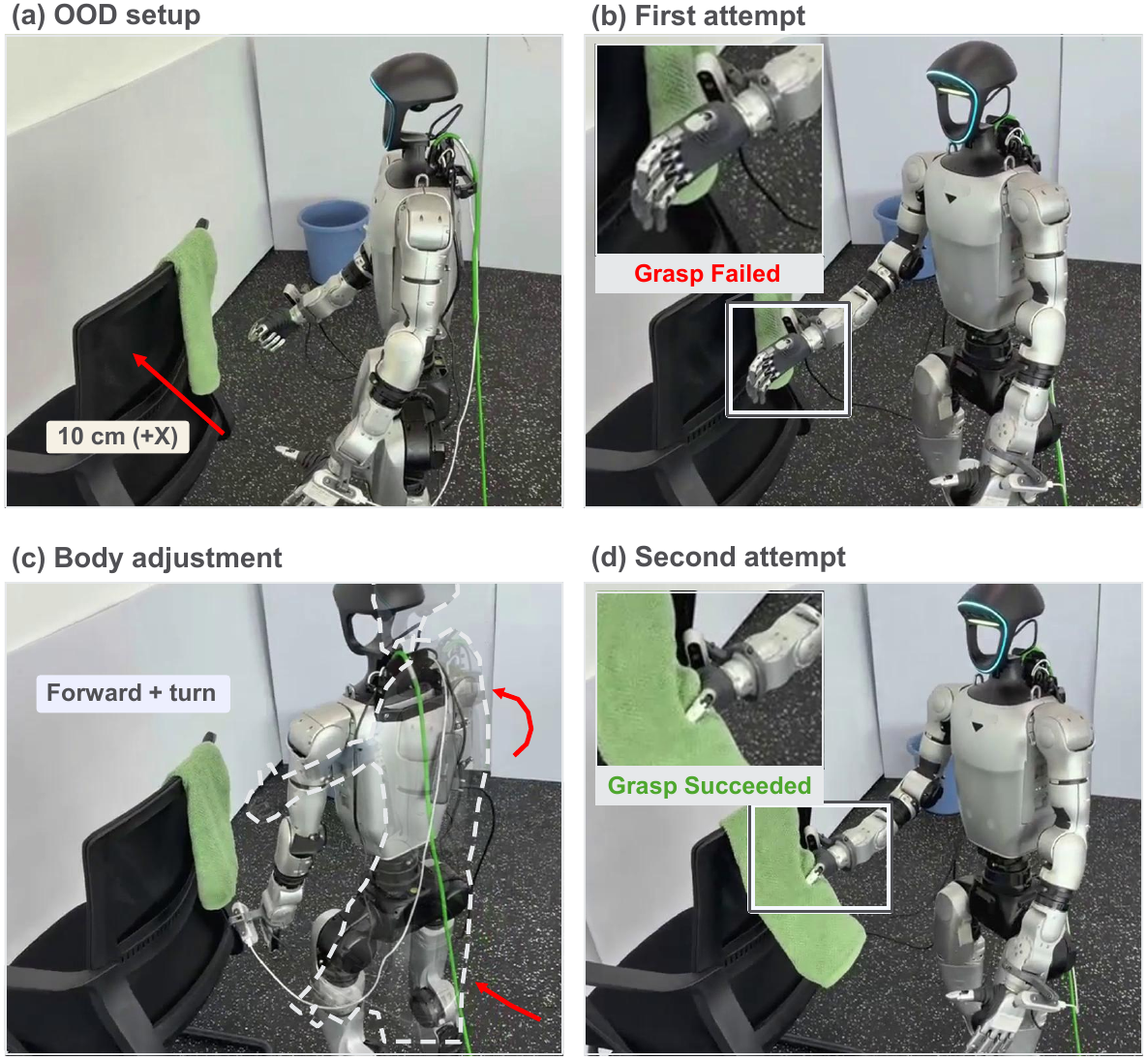}
 \makeatletter
 \long\def\@makecaption#1#2{%
   \@IEEEfigurecaptionsepspace
   \parbox[t]{\hsize}{\footnotesize\noindent
     \mbox{#1.\hspace{0.5em}}#2}%
 }
 \makeatother
 \caption{\textbf{Whole-body coordination generalization.} Under a 10 cm displacement of the towel support, \ours{} recovers from a failed grasp through forward repositioning and body rotation.}
 \label{fig:coordination_generalization}
\end{figure}

\textit{Metrics.} We adopt three quantitative metrics. (1) \textit{Task Success Rate (TSR)} is the percentage of trials completing all task objectives within the evaluation horizon. (2) \textit{Task Progress (TP)}~\cite{omega} measures normalized progress along the ordered task stages before the first failure or irreversible deviation. (3) \textit{Cross-WBC Variance} is the population variance of mean TSR across different downstream controllers, measured in pp$^2$, with lower values indicating greater robustness. All metrics use 20 independent trials for each combination of method and task. TSR and TP are averaged over trials, while cross-WBC variance is computed from the resulting controller-wise TSRs.

\subsection{Simulation Experiments}
\textit{Overall performance.} Table~\ref{tab:simulation_quantitative_ablation} reports performance across six tasks and three SIMPLE perturbation levels. \ours{} achieves the highest overall TSR of 91.9\%, exceeding the WAM baselines DreamZero, DreamZero-PT, and Cosmos-3 by 26.9, 18.3, and 5.6 percentage points, respectively. The advantage over DreamZero-PT, which uses the same post-training data, indicates that additional humanoid supervision alone does not explain the gains. Compared with Cosmos-3, improvements are larger on coordination-intensive PickBetweenTables and MoveBendPick (10.0 and 11.7 points), while the two methods achieve comparable task-averaged TSRs on Handover and TabletopGrasp. \ours{} also outperforms the VLA baselines $\Psi_0$ (82.2\%) and GR00T N1.6 (47.5\%), exceeding the highest VLA baseline TSR by 9.7 percentage points.

\textit{Cross-WBC robustness.} We additionally evaluate each method across tasks using AMO~\cite{amo}, SONIC~\cite{sonic}, and GEAR WBC~\cite{gearwbc}, with separate task-specific fine-tuning for each controller. Following Table~\ref{tab:uwbc_slot_specification}, SONIC uses joint-space slots $[16,46)$ for lower-body/waist references. AMO maps planar velocity, heading, absolute height, and torso orientation to $[0,2)$, $[6,8)$, $10$, and $[13,16)$, respectively, with its turning flag in residual slots. GEAR WBC uses the same velocity, height, and orientation slots, with its derived yaw rate in slot $5$. For each target WBC, fine-tuning activates only fields corresponding to controller commands available in the demonstrations and masks all other registered fields. Figure~\ref{fig:controllers} shows that \ours{} achieves the highest TSR averaged across the three controllers (89.2\%) and the smallest cross-controller variance (10.5 pp$^2$), compared with 80.2\% and 35.0 pp$^2$ for Cosmos-3. For baselines initially trained with a single WBC such as SONIC, adaptation requires learning new command conventions without a shared prior across interfaces. This provides a plausible explanation for their performance degradation and greater sensitivity to WBC choice. By separating internal whole-body intent from controller-specific command conventions, UWBC supports more compatible and stable adaptation across the evaluated WBC interfaces.

\Needspace{3\baselineskip}
\textit{Downstream fine-tuning efficiency.} Figure~\ref{fig:scaling} compares TSR on four simulation tasks using 50/100/200/300 task-specific fine-tuning demonstrations per task; 100 demonstrations is the main simulation budget. \ours{} achieves its largest advantage with few task-specific demonstrations. The results support effective reuse of pre-trained world--action priors for humanoid loco-manipulation through structured WBC grounding and coordination.

\subsection{Real-World Experiments}
We evaluate \ours{} and DreamZero on eight real-world tasks (Fig.~\ref{fig:real_tasks}) under ID and OOD conditions. ID follows the training distribution, whereas OOD changes object configurations and language instructions.

\textit{ID performance.} As shown in Table~\ref{tab:real_world_id_ood_tsr}, \ours{} achieves a mean TSR of 81.3\%, compared with 57.5\% for DreamZero. The gap is small on manipulation-dominant tasks such as TableCleanup (90\% versus 85\%). This suggests that pre-trained manipulation priors already support effective execution when targets lie within the arm workspace and require little base or torso adjustment. As tasks involve more body subsystems and coordination stages, baseline performance declines, suggesting that treating manipulation and WBC commands as homogeneous action dimensions may inadequately capture complex whole-body coordination behavior. In contrast, \ours{} maintains strong performance, with gains of 35 points on BasketCarry and DoorEntry and 40 on BoxTransfer. These gains are consistent with the benefits of structured WBC grounding and manipulation-informed coordination on tasks requiring substantial body reconfiguration.

\Needspace{4\baselineskip}
\textit{OOD generalization.} \ours{} maintains a mean TSR of 68.8\%, compared with 40.0\% for DreamZero, with a smaller ID-to-OOD drop (12.5 versus 17.5 points). The advantage reaches 45 points on DoorEntry, which requires substantial whole-body adjustment. Figure~\ref{fig:progress} shows mean ID/OOD task progress of 0.92/0.82 for \ours{} versus 0.73/0.59 for DreamZero, with strong performance on manipulation-dominant tasks and larger gains on coordination-intensive tasks. These results suggest that \ours{} preserves task-level manipulation performance while adapting its execution to new whole-body coordination requirements under distribution shift. Moreover, we observe notable whole-body coordination generalization in OOD settings, as shown in Fig.~\ref{fig:coordination_generalization}. In TowelPlace, displacing the towel support 10 cm farther from the robot puts the towel beyond the initial arm workspace. After one unsuccessful local reaching attempt, \ours{} takes a short forward step and adjusts its torso to extend its reach, successfully grasping the towel on the next attempt. This recovery sequence is absent from our demonstrations. We interpret it as emergent behavior arising from pre-trained WAM manipulation priors coupled with enhanced whole-body coordination.

\subsection{Ablation Studies}
We ablate CASA, UWBC, and structured action factorization (SAF) across six simulation tasks (Table~\ref{tab:simulation_quantitative_ablation}). Removing CASA retains the manipulation and UWBC streams but restores native self-attention, reducing mean TSR from 91.9\% to 86.9\% (5.0 points). The reduction is larger on PickBetweenTables (8.3 points) and MoveBendPick (6.7 points) than on Handover (1.7 points), consistent with the benefit of state-dependent directional coordination on tasks requiring greater whole-body adjustment. Replacing UWBC with native SONIC commands while retaining the WBC pathway and supervision reduces TSR to 84.7\% (7.2 points). This suggests that explicit physical semantics improve controller grounding even when a distinct WBC pathway is available. Finally, removing SAF replaces the manipulation and UWBC streams with a capacity-matched monolithic stream under the same data and optimization budget, reducing TSR to 80.8\% (11.1 points). The largest reduction supports preserving the pre-trained manipulation pathway and keeping WBC commands separately addressable within joint generation.

\section{Conclusion}
We present \ours{}, a WBC-grounded world action model for coordinated humanoid loco-manipulation. \ours{} preserves pre-trained world--action priors and grounds heterogeneous whole-body commands through UWBC. Coordination-aware self-attention further enhances state-dependent manipulation-to-UWBC interaction. Experiments in simulation and on real robots demonstrate improved task performance, OOD generalization, and cross-WBC robustness while maintaining strong performance on manipulation-dominant tasks. Data-scaling and ablation results further support the value of structured prior reuse. More broadly, our results suggest that scalable humanoid whole-body intelligence can be built by generalizing reusable world--action priors through structured WBC grounding and coordination, rather than relearning whole-body behavior from scratch.

\bibliographystyle{IEEEtran}
\bibliography{references}
\end{document}